\documentclass[10pt,journal,compsoc]{IEEEtran}
\PassOptionsToPackage{hyphens}{url}\usepackage{hyperref}
\usepackage{graphicx}
\usepackage{blindtext}
\usepackage{amsmath}
\usepackage{amsfonts}
\usepackage{booktabs}
\usepackage{multirow}
\usepackage{makecell}
\usepackage[scientific-notation=true]{siunitx}
\usepackage{pgf}
\usepackage{pgfpages}
\usepackage{hyperref} 
\usepackage{arydshln}
\usepackage{cite}

\renewcommand{\textdownarrow}{$\downarrow$}
\renewcommand{\textuparrow}{$\uparrow$}

\newcommand{\mathboldit}[1]{\boldsymbol{#1}}
\usepackage{todonotes}

\begin{document}

\title{Multi-Camera Trajectory Forecasting with Trajectory Tensors}
%
%
%
%

\author{Olly~Styles,
        Tanaya~Guha,
        and~Victor~Sanchez
\IEEEcompsocitemizethanks{\IEEEcompsocthanksitem The authors are with the Department of Computer Science, University of Warwick, United Kingdom.  Contact email: o.c.styles@warwick.ac.uk}
\thanks{Manuscript received November 13, 2020; revised May 12, 2021.}}

%
%

\markboth{Journal of \LaTeX\ Class Files,~Vol.~14, No.~8, August~2015}%
{Shell \MakeLowercase{\textit{et al.}}: Bare Demo of IEEEtran.cls for Computer Society Journals}
%



\IEEEtitleabstractindextext{%
\begin{abstract}
We introduce the problem of multi-camera trajectory forecasting (MCTF), which involves predicting the trajectory of a moving object across a network of cameras. While multi-camera setups are widespread for applications such as surveillance and traffic monitoring, existing trajectory forecasting methods typically focus on single-camera trajectory forecasting (SCTF), limiting their use for such applications. Furthermore, using a single camera limits the field-of-view available, making long-term trajectory forecasting impossible. We address these shortcomings of SCTF by developing an MCTF framework that simultaneously uses all estimated relative object locations from several viewpoints and predicts the object's future location in all possible viewpoints. Our framework follows a \emph{Which-When-Where} approach that predicts in \emph{which} camera(s) the objects appear and \emph{when} and \emph{where} within the camera views they appear. To this end, we propose the concept of \emph{trajectory tensors}: a new technique to encode trajectories across multiple camera views and the associated uncertainties. We develop several encoder-decoder MCTF models for trajectory tensors and present extensive experiments on our own database (comprising 600 hours of video data from 15 camera views) created particularly for the MCTF task. Results show that our trajectory tensor models outperform coordinate trajectory-based MCTF models and existing SCTF methods adapted for MCTF. Code is available from: \url{https://github.com/olly-styles/Trajectory-Tensors}
\end{abstract}

\begin{IEEEkeywords}
Trajectory forecasting, multi-camera tracking, person re-identification, multi-camera trajectory forecasting.
\end{IEEEkeywords}}

\maketitle

\IEEEdisplaynontitleabstractindextext

%
\IEEEpeerreviewmaketitle

\begin{figure}[t]
\centering
\includegraphics[width=\linewidth]{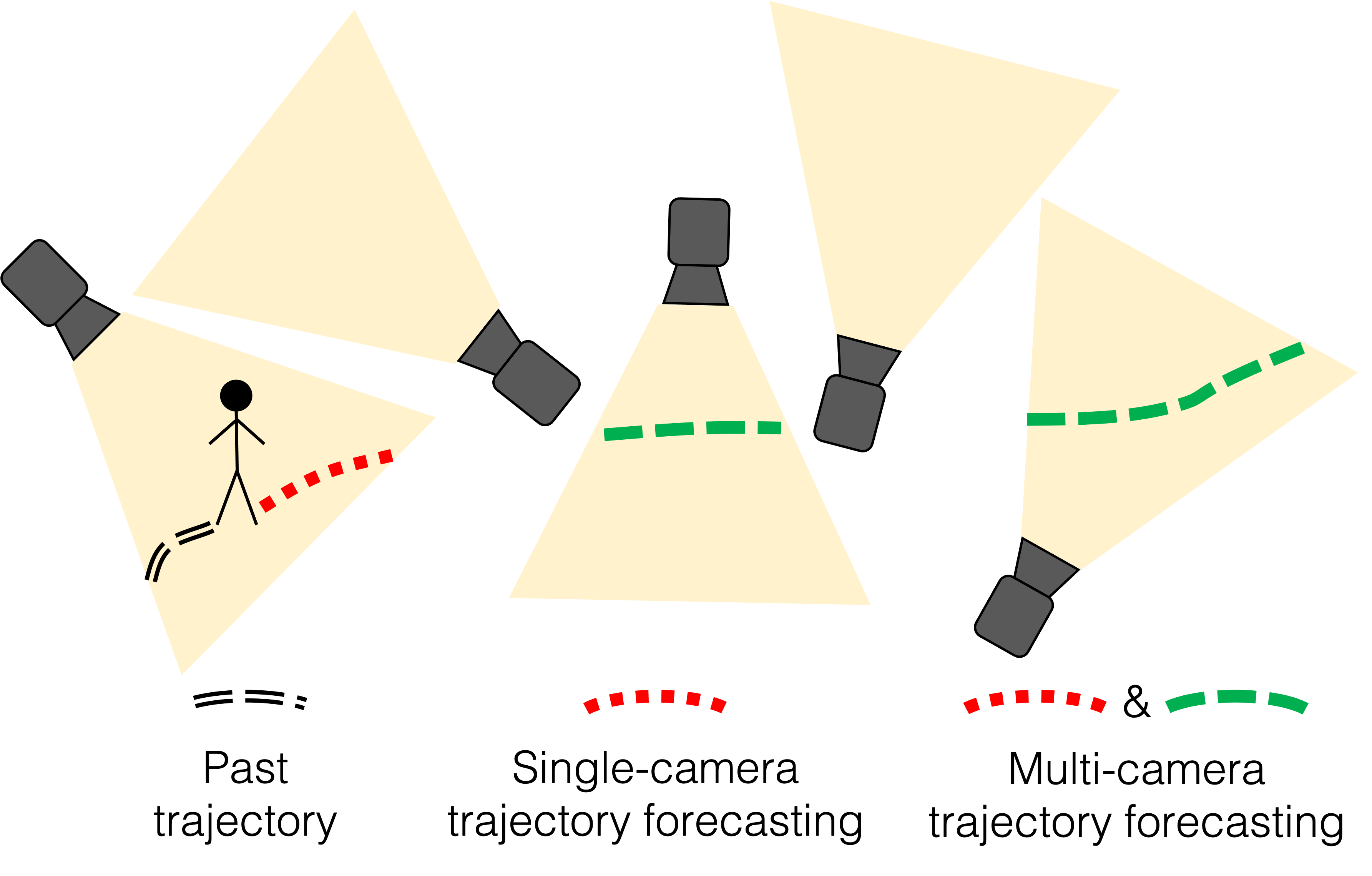}
\caption{\textbf{Multi-camera trajectory forecasting}. We introduce a novel formulation of the trajectory forecasting task which considers multiple camera views. Companion video: \url{https://youtu.be/IjlNEvKQ634}} 
\label{fig:mctf}
\end{figure}

\IEEEraisesectionheading{\section{Introduction}\label{sec:introduction}}
\IEEEPARstart{P}{redicting} the future trajectory of objects in videos is a challenging problem with multiple application domains such as intelligent surveillance \cite{surveillance-survey}, autonomous driving \cite{desire}, person re-identification (RE-ID) \cite{spatio-temporal-reid}, and traffic monitoring \cite{cityflow}. Existing works in this area focus on single-camera trajectory forecasting (SCTF), i.e., predicting a future trajectory of an object within the same camera view in which the object is observed \cite{behavior-cnn,social-lstm,social-gan,sophie,desire,fpl,mof, mantra}. A critical drawback of the single-camera settings is that models cannot anticipate when new objects will enter the scene \cite{density-forecasting} as they are limited to the data from a single camera. Furthermore, SCTF methods are only suitable for short-term trajectory forecasting, typically 1 to 5 seconds \cite{social-lstm,sophie,social-gan,fpl,mof, mantra}, due to the limited field-of-view. The constraint of a single camera viewpoint must be removed to overcome these issues. To this end, we introduce \textit{multi-camera trajectory forecasting} (MCTF) - a new framework within trajectory prediction. Given the information about an object's location in one or more camera views, we want to predict its future location across a camera network in all possible camera views. Fig.~\ref{fig:mctf} presents an overview of the MCTF framework. 


Tracking an object-of-interest across a large camera network requires simultaneously running state-of-the-art algorithms for object detection, tracking, and RE-ID. These algorithms can be computationally demanding, particularly for large camera networks. Processing videos at a lower image resolution or frame-rate may reduce the computational demands, but this often results in missed detections \cite{yolo-9000}. Alternatively, we may choose to monitor only a subset of the cameras in a network, resulting in missed detections. A successful MCTF model can mitigate this issue by preempting an object's location in a distributed camera network, allowing the system to monitor fewer cameras through an intelligent selection technique. This is distinct from previous works that use trajectory information for person RE-ID \cite{spatio-temporal-reid} or vehicle tracking \cite{multi-camera-vehicle-tracking} which are \textit{reactive} to observations \emph{after} an object has been observed in multiple camera views. Our proposed MCTF framework is \textit{proactive} - it predicts the future location of an object \emph{before} it enters the camera view.

We present a deep encoder-decoder approach to MCTF that introduces the idea of \emph{trajectory tensors} - a new technique to encode trajectories across multiple camera views and the associated uncertainty in future locations. Trajectory tensors are an attractive alternative to the coordinate-based trajectory, which is the de facto representation in existing trajectory forecasting works \cite{social-lstm,sophie,social-gan,fpl,mantra,mof}, with only a few recent exceptions \cite{garden-of-forking}. Coordinate trajectories represent the historical and predicted future locations of an object as a sequence of coordinates. The representation considers coordinates in the image space in a single-camera view or world coordinates if a projection is available. In contrast to the coordinate approach, proposed trajectory tensors divide viewpoints from several cameras into grid cells with values indicating an object's presence or absence. This representation enables us to intuitively model diverse future locations, their associated uncertainty, and object locations in an arbitrary number of viewpoints. Trajectory tensors also offer easy to interpret results that can be visualized easily. 

Given an object tracklet in one or more camera view(s), our MCTF framework comprises the following three tasks: (i) In \textit{which} cameras will the object appear next? (ii) \textit{When} will the object appear in those cameras? (iii) \textit{Where} will the object appear in the identified camera views? We use an object tracklet of 2 seconds to predict 12 seconds into the future. Owing to the wide body of complementary literature on pedestrian detection \cite{detection-survey} and person RE-ID \cite{re-id-survey}, we focus on pedestrians for our MCTF task. Nevertheless, the task, along with our proposed data representation and model, can be easily generalized to any moving object. 

MCTF and the Warwick-NTU Multi-camera Forecasting (WNMF) dataset were first introduced in our earlier work \cite{mctf}. This work extends our previous work as follows:

\begin{itemize}
\item We introduce three problem formulations within the MCTF framework: \textit{which}, \textit{when}, and \textit{where} (Section \ref{sec:mctf}). Each task predicts trajectories with more granularity than the previous. 
\item We propose trajectory tensors, a new data representation for multi-camera trajectories (Section \ref{sec:data_representation}). Trajectory tensors overcome many of the shortcomings of coordinate trajectories for MCTF. 
\item We present a deep encoder-decoder for MCTF (Section \ref{sec:models}). Our novel approach, based on trajectory tensors, outperforms several relevant baselines (Section \ref{sec:evaluation}).
\end{itemize}

\section{Related work} \label{sec:related-work}

To the best of our knowledge, our work is the first to consider trajectory forecasting in a non-overlapping multi-camera setting. Therefore, we review literature in the most related fields.

\vspace{0.5mm}\noindent\textbf{Single-camera trajectory forecasting.} Trajectory forecasting has seen considerable attention in recent years, generally focusing on forecasting pedestrian trajectories from a birds-eye viewpoint \cite{eth,ucy}. Many approaches use recurrent models such as long short-term memory (LSTM) networks and focus on extracting features to encode social norms such as avoiding collisions with others, and group movements \cite{social-lstm,social-gan,sophie}. Other works also consider environmental constraints \cite{deepcontextmap}. A comparatively small number of works consider visual features for trajectory forecasting, such as those extracted from human pose \cite{fpl,peeking-into-the-future} or optical flow \cite{mof} from non-nadir viewpoints. Models are often optimized using loss functions such as mean squared error (MSE) \cite{fpl,mof,car-net}. As the distribution of future trajectories is multi-modal, using the MSE loss heavily penalizes reasonable but incorrect forecasts, such as turning right rather than left at an intersection. Rather than using MSE, some works use an adversarial loss to train models capable of generating multiple possible future paths using generative adversarial networks (GANs) \cite{social-gan,sophie} or variational auto-encoders (VAEs) \cite{destination-not-journey,desire}. However, evaluating multi-output models can be challenging, as typically, only a single ground truth trajectory exists despite multiple plausible futures. Liang et al. overcome this issue by generating a simulated dataset where each trajectory has multiple futures \cite{garden-of-forking}. Note that all of the works mentioned above consider only a single camera viewpoint at a time.
\vspace{0.5mm}\\\noindent\textbf{Person RE-ID.} Person RE-ID is the task of identifying an individual in a set of gallery images given a probe image, often in a multi-camera setting. Substantial progress has been made in this area over recent years. 
For example, on the commonly used Market \cite{marketdataset} benchmark, rank 1 person RE-ID performance (i.e. correct RE-ID given 1 guess only) has improved from 47.3\% in 2015 \cite{marketdataset} to 96.8\% in 2020 \cite{aaai-2020-reid}. Most work on person RE-ID has focused on image-level matching, and current state-of-the-art methods \cite{auto-reid,bagoftricks} exploit this visual cue without other sources of information.  However, image-level similarity matching is just a single component of a comprehensive RE-ID system. Persons must first be detected and tracked before matching, which can be computationally expensive to run simultaneously on a large network of cameras. 
Incorporating trajectory information for RE-ID in a multi-camera setting has seen comparatively little attention, in part due to a lack of publicly available datasets. One notable exception is the recent work of Wang et al. \cite{spatio-temporal-reid}. The authors demonstrate that by retrospectively utilizing trajectory information and visual features, their approach can attain state-of-the-art RE-ID results on the prevalent Duke-MCMT benchmark dataset \cite{dukedataset}. The Duke-MCMT dataset is no longer available to the research community \cite{duke-dataset-removed}.
\vspace{0.5mm}\\\noindent\textbf{Multi-camera surveillance.} A typical automated multi-camera surveillance system consists of detection, tracking, and RE-ID components to monitor objects-of-interest \cite{distributed-reid,scaling-analytics,lab-to-real-world}. Scaling such systems to large camera networks can be challenging due to computational demands. Jain et al. \cite{scaling-analytics} study the impact of scaling multi-camera tracking to large networks and show that filtering the search space to high traffic areas can considerably reduce the search space at only a small cost in recall. In addition to using the traffic level in each area, trajectory information has also been used for both RE-ID \cite{spatio-temporal-reid} and multi-camera tracking \cite{multi-camera-vehicle-tracking}. Using trajectories for these tasks can supplement existing appearance-based models by providing a second source of information for matching objects across camera views. However, previous works \cite{spatio-temporal-reid,multi-camera-vehicle-tracking,scaling-analytics,lab-to-real-world,distributed-reid} reactively use trajectory information, i.e., the object must have already been observed in multiple camera views. Alternatively, Alahi et al. \cite{cvpr-mctf} predict the origin and destination of pedestrians using a dense multi-camera setup. The authors use nadir viewpoints and depth sensors to map tracks to the world coordinate space. Our method, in contrast, predicts trajectories in the image space. Our method is therefore applicable to a typical video surveillance setup with non-overlapping views and standard cameras without depth sensors.

\section{The MCTF framework} \label{sec:mctf}

Consider a typical multi-camera surveillance setup where a set of $k$ cameras $\mathcal{C}=\{c_i\}_{i=1}^k$ are mounted overhead monitoring objects of interest. Given an object's bounding boxes from the previous $n$ timesteps up to the current timestep $t$, $B_{(t-n:t)} = \{B_{(t-n)}, \cdots B_{(t-1)}, B_{(t)}\}$, we propose a hierarchy of MCTF problem formulations.

\vspace{1mm}
\noindent\textbf{In \textit{which} camera(s) will the object appear in the future?} Given $B_{(t-n:t)}$, our task is to identify a subset of $\mathcal{C}$ in which the object may appear at any future timestep, up to a maximum of $m$ timesteps. We cast this as a multi-class multi-label classification problem, where an object can appear in any number of cameras in $\mathcal{C}$. Our goal is to estimate the probability of appearance of the object $P_a(c_i|B_{(t-n:t)})$ for each camera $c_i \in \mathcal{C}$, where a positive class is a camera in which the object re-appears. The output is a vector $[P_a(c_1|B_{(t-n:t)}), \ldots, P_a(c_k|B_{(t-n:t)})]$ of length $k$. 

\vspace{1mm}
\noindent\textbf{\textit{When} will the object appear?} The task here is to predict when the object will reappear within the next $m$ timesteps in a given camera. Similar to the \textit{Which} problem, we also formulate this as a multi-class multi-label problem, where we compute the joint probability of appearance of the object $P_a(c_i,t_j|B_{(t-n:t)})$ for each camera $c_i \in \mathcal{C}$ at each timestep $t_j$ with $j = 1, \cdots, m$. The output is a matrix $\big[\begin{smallmatrix}
  P_a(c_1,t_1|B_{(t-n:t)}) & \cdots \\
  \cdots & P_a(c_k,t_m|B_{(t-n:t)})
\end{smallmatrix}\big]$ of dimension $k\times m$. 

\vspace{1mm}
\noindent\textbf{\textit{Where} will the object appear?} This task aims at spatially localizing the object within a camera view in addition to \textit{which} and \textit{when}. To this end, we divide each camera view into a $w \times h$ grid and predict a probability of appearance score $P_a(c_i,t_j,g_{xy}|B_{(t-n:t)})$ for each grid cell $g_{xy}$, where $x = 1,\cdots, w$ and  $y=1,\cdots, h$. The output is a tensor $\mathbf{Z}$ of dimension ${k \times m \times w \times h}$ containing the probability of appearance scores $P_a(c_i,t_j,g_{xy}|B_{(t-n:t)})$ for each camera $c_i \in \mathcal{C}$, timestep $t_j$ with $j = 1, \cdots, m$, and grid cell $g_{xy}$ with $x = 1, \cdots, w$ and $y = 1, \cdots, h$. 

It is important to note that this work focuses on modeling target locations in a multi-camera setting where accurate geometry information is not available. We predict the location of a single target unless otherwise specified.



\section{Proposed approach} \label{sec:representation-and-model}

This section introduces our data representation technique, i.e., trajectory tensors, our models, and evaluation strategy for MCTF.

\subsection{Trajectory Tensors} \label{sec:data_representation}

Existing works represent trajectories using a coordinate approach \cite{social-lstm,sophie,social-gan}. Coordinate trajectories are $(x,y)_t$ vectors in the image or world space, representing the location of an object at a particular timestep $t$. Sometimes, the width ($w$) and height ($h$) of the object may be also included in the representation, i.e., $(x,y,w,h)_t$ \cite{brian-egocentric,mof}.

There are three critical drawbacks to a coordinate trajectory representation: \\(i) Standard coordinate-based approaches generally do not define a null trajectory. This representation can be problematic in real-world scenarios where object coordinates can become unavailable, such as when the object is occluded, or the detection algorithm fails. It is particularly problematic in a multi-camera scenario where objects are not visible in all camera views simultaneously. \\(ii) Coordinate trajectories can only represent a trajectory as viewed from a single camera. Due to the null trajectory problem, coordinates cannot be easily generalized to multiple cameras, 
unless all objects are simultaneously visible in all cameras in $\mathcal{C}$, or separate models are created for each camera. Trajectories can instead be mapped to the world coordinate space to overcome this issue. However, this mapping requires accurate measurements of the camera locations and intrinsic parameters which are not always available. This work assumes that such information is not available.
\\(iii) Finally, coordinate trajectories do not inherently represent uncertainty, which is intrinsic to the trajectory forecasting task. The space of future trajectories is multi-modal, e.g., an object can travel either left or right at a junction. Existing works address this issue by using generative models to simulate multiple futures \cite{social-gan, garden-of-forking}, from which a probability distribution can be created. Our approach, in contrast, intuitively models uncertainty in one shot. 

\begin{figure}[t]
\centering
\includegraphics[width=0.9\linewidth]{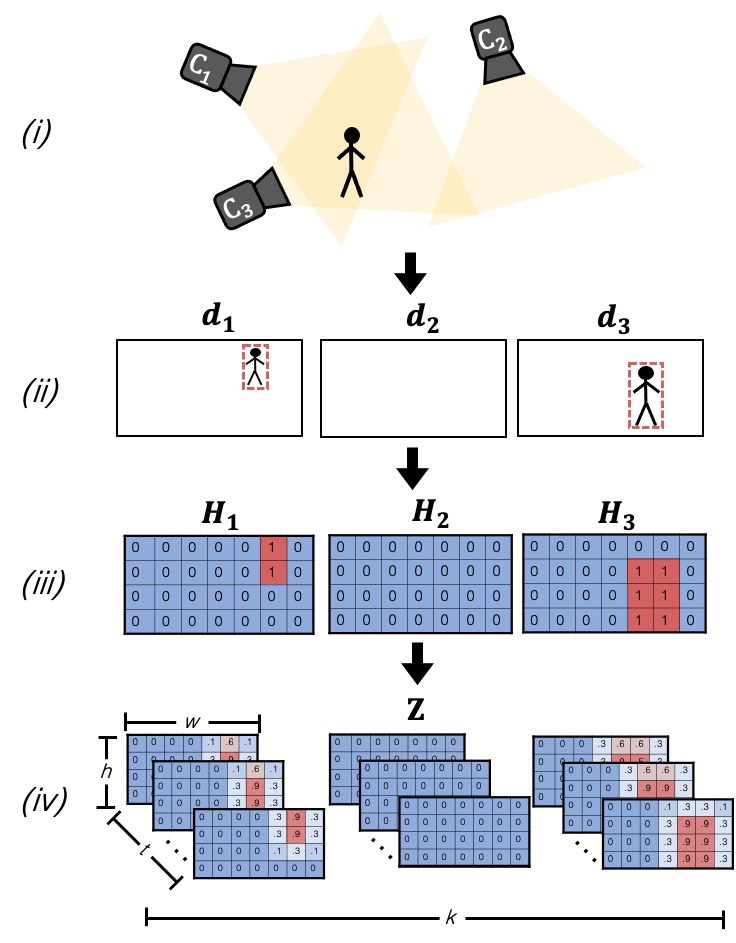}
\vspace{-2mm}
\caption{\textbf{Trajectory tensors.} Our proposed trajectory tensors are an intuitive data representation capable of representing object trajectories in multiple camera views, null trajectories, and associated uncertainty.}
\label{fig:tensors}
\end{figure}

To overcome the shortcomings of coordinate trajectory representation, we introduce the idea of \emph{trajectory tensors} - a novel technique for compact representation of multi-camera trajectories. As shown in Fig. \ref{fig:tensors}, trajectory tensors are constructed in four steps:

\noindent\textit{(i)} We consider a set of cameras $\mathcal{C} = \{{c_i}\}_{i=1}^{k}$, where an object of interest may appear in any number of the cameras in $\mathcal{C}$. 

\noindent\textit{(ii)} Each camera $c_i$ has an associated detection $\mathbf{d_i}$ representing the object bounding box, if present. 

\noindent\textit{(iii)} We convert each $\mathbf{d_i}$ into a heatmap $\mathboldit{H_i}$ which is an alternative representation of the object's location. A heatmap is a matrix of size $w \times h$, where each entry is a binary value indicating the presence or absence of the object in this grid cell. A single object may span any number of grid cells, depending on its size. 

\noindent\textit{(iv)} Finally, the heatmaps $\mathboldit{H_i}$ for each of the $k$ cameras are stacked along the camera dimension, and computed for $t$ timesteps. We also smooth each heatmap using a Gaussian kernel. The result is a trajectory tensor, $\mathbf{Z}$, of dimension $k \times t \times w \times h$, where each entry is a value between 0 and 1. $\mathbf{Z}$ represents the trajectory of an object in multiple camera views simultaneously and can be used to represent both past (inputs) and future (predicted) object locations. Encoding object locations as trajectory tensors allows us to elegantly represent objects in multiple cameras. Trajectory tensors enable the representation of null trajectories rather than discarding this data or using placeholder values.

Trajectory tensors share similarities with the grid-cell representation proposed recently by Liang et al. \cite{garden-of-forking}; however, our proposed encoding includes multiple camera views. Besides, objects represented using trajectory tensors may span multiple grid cells, which allows us to account for variability in object scale. Object scale is not considered in most SCTF works \cite{garden-of-forking,social-lstm,social-gan,sophie}, but is a critical component in our framework. 



\subsection{MCTF models} \label{sec:models}

\begin{figure*}[t]
\centering
\includegraphics[width=0.9\linewidth]{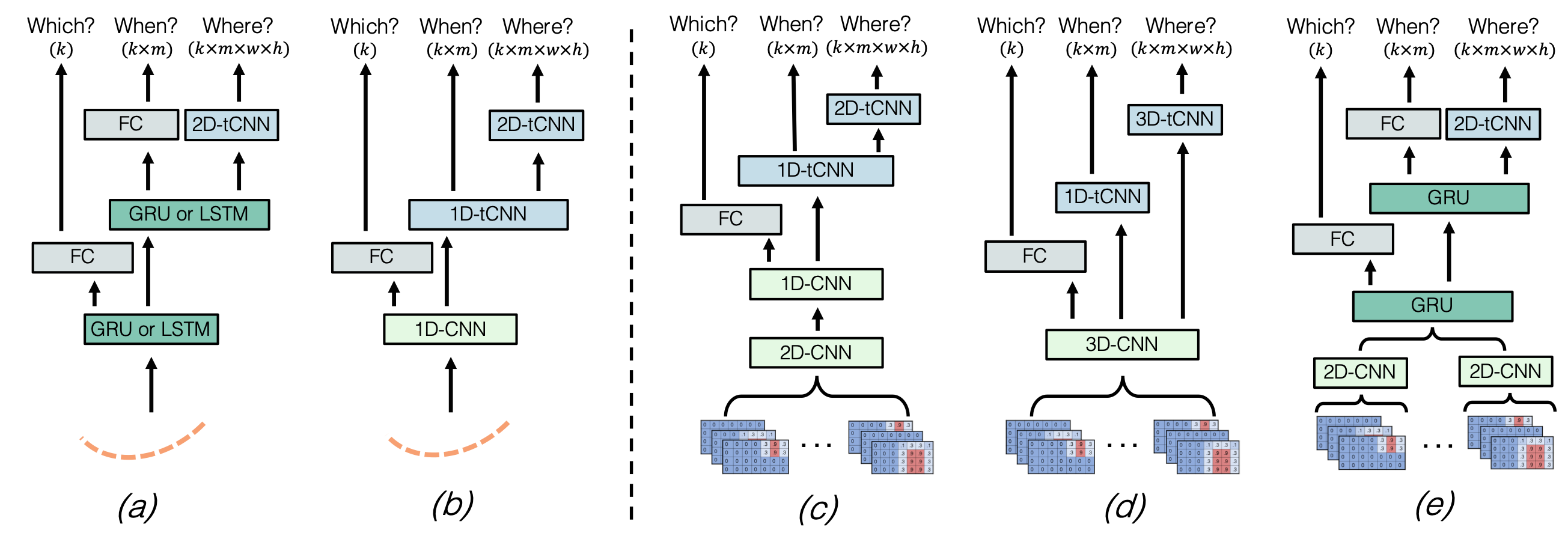}
\vspace{-4mm}
\caption{\textbf{MCTF models.} We introduce 2 coordinate-trajectory based (left) and 3 trajectory-tensor based (right) approaches for MCTF. \textit{(a)} A recurrent encoder-decoder adapted from \cite{mof}. \textit{(b)} A 1D-CNN adapted from \cite{fpl}. \textit{(c)} A CNN approach with separated layers for spatial and temporal feature extraction. \textit{(d)} A CNN approach with 3D convolutions for extracting spatial and temporal features simultaneously. \textit{(e)} A hybrid CNN-GRU approach which uses a CNN to extract spatial features which are passed to an GRU for extracting temporal features. Note that for coordinate trajectory models, a separate model is created for each camera. In contrast, trajectory tensor models use a single unified model for all cameras. Each model is trained with either the \textit{which}, \textit{when}, or \textit{where} MCTF formulation.}
\label{fig:mctf_models}
\end{figure*}

Existing trajectory forecasting methods such as Social-LSTM \cite{social-lstm}, Social-GAN \cite{social-gan}, and SoPhie \cite{sophie} are designed for SCTF using datasets with birds-eye view cameras \cite{ucy,eth} that do not include object scale. Although the methods proposed in \cite{fpl} and \cite{mof} forecast object scale in addition to location, they are also designed for SCTF; hence direct comparison for MCTF is not possible. To compare with these existing works, we adapt the methods proposed in \cite{fpl} and \cite{mof} to our MCTF framework. These models, along with new approaches based on trajectory tensors, are summarized in Fig. \ref{fig:mctf_models}. Each neural network-based model is comprised of a combination of fully-connected, recurrent, and convolutional layers. Inspired by fully-convolutional networks for semantic segmentation \cite{fully_conv}, we use transposed convolutional layers when predicting trajectory tensors as targets. Due to the large number of neural network models introduced in this section, we introduce each briefly and provide full details of the architectures needed to reproduce our results in the supplementary materials. 

\subsubsection*{Coordinate trajectory approaches} \label{sec:coordinate_trajectories}

Here we present our coordinate trajectory approaches which are shown in Fig. \ref{fig:mctf_models} (a) and (b). Due to the lack of data representations for coordinate trajectories in multiple camera views, our coordinate trajectory approaches each use a separate model for each camera, resulting in $k$ models.

\vspace{0.5mm}
\noindent\textbf{GRU.} Recurrent networks have been a prevalent approach for trajectory forecasting in a single camera. We use an adapted version of STED \cite{mof}, a method for SCTF which uses gated recurrent units (GRUs) with an
encoder-decoder architecture. The model proposed in the
original work uses two encoders, one for bounding box
coordinates, and another which uses a convolutional neural network (CNN) to extract motion information from optical flow. We use only the bounding box encoder for a fair comparison as other methods do not use visual features. For the \textit{which} task, we use a fully-connected classification layer. For the \textit{when} task, we use another GRU as a decoder with 128 hidden units followed by a fully connected classification layer. For the \textit{where} task, the decoder GRU is followed by 2D transposed convolutional (tCNN) layers for spatial upsampling.

\vspace{0.5mm}
\noindent\textbf{LSTM.} Our LSTM model is the same as our GRU, with an alternative recurrent unit for both encoder and decoder. 

\vspace{0.5mm}
\noindent\textbf{1D-CNN.} 1D-CNNs have received some attention as an alternative to the de facto recurrent models for tasks involving time series \cite{time-series-review}, including trajectory forecasting \cite{fpl}. We use the encoder architecture proposed by Yagi et al. \cite{fpl}, with modified decoders adapted for the MCTF formulation. For the \textit{which} task, we use a fully-connected classification layer. For the \textit{when} task, we use 1D transposed convolutional layers as a decoder with four layers. For the \textit{where} task, the 1D transposed convolutional layers are followed by 2D transposed convolutional layers for spatial upsampling. Similarly to STED, we use only the trajectory feature extractor for a fair comparison with other methods. 

\subsubsection*{Trajectory tensor approaches}

Here we present our trajectory tensor approaches which are shown in Fig. \ref{fig:mctf_models} (c), (d), and (e). Our proposed representation enables efficient modeling of object trajectories across multiple camera views. Therefore, each approach consists of a single unified model for all cameras, which is less cumbersome and more scalable than the multi-model approach using coordinate trajectories.


\vspace{0.5mm}
\noindent\textbf{3D-CNN.} We use a 3D-CNN with four layers for spatio-temporal feature extraction. 3D convolutions can simultaneously extract spatial and temporal features and have seen some success for tasks such as action recognition \cite{can-3d-follow-2d}.

\vspace{0.5mm}
\noindent\textbf{2D-1D-CNN.} We use a CNN consisting of three 2D convolutional and pooling layers for spatial feature extraction, followed by three 1D convolutional and pooling layers for temporal feature extraction. Using separate 2D and 1D convolutional layers rather than 3D layers reduces the number of network parameters and is inspired by existing models for video classification \cite{2d-1d-video-classification}.

\vspace{0.5mm}
\noindent\textbf{CNN-GRU.} We train a convolutional auto-encoder to reduce a trajectory tensor at a single timestep (size $k \times w \times h$) to a feature vector of size 128. This encoding is then used as the input to a GRU which predicts future feature vectors, which are decoded as shown in Fig. \ref{fig:mctf_models}~(e). The auto-encoder is first pre-trained until convergence and then trained end-to-end with the GRU. This architecture is inspired by \cite{forecasting-instance-segmentation}, which uses a similar strategy of forecasting future convolutional features for predicting future instance segmentation maps.

\subsection{Evaluation strategy}

Due to the high levels of uncertainty and the multi-modal nature of the MCTF, traditional SCTF metrics such as average and final displacement errors are not well-suited to MCTF without modification. Some works generate multiple trajectories and select the one most similar to the ground truth \cite{social-gan}; however, this evaluation method is optimistic as performance comparable to sophisticated methods can be obtained using a simple constant velocity model that generates trajectories with high variance \cite{constant-velocity}. We instead compute the average precision (AP) for all problem formulations. We plot precision-recall curves rather than ROC curves. We find ROC to be an overly-optimistic metric due to the considerable class imbalance between the positive (object presence) and negative (no object presence) classes. We define $AP_{which}$, $AP_{when}$, and $AP_{where}$ for the three problem definitions respectively.

The AP metrics provide a holistic interpretation of model performance but are not easy to interpret. Therefore, we propose two new metrics for MCTF evaluation, the soft-intersection-over-union ($SIOU$) for the when and where problem formulations. The $SIOU$ for evaluating the \textit{when} formulation is as follows:

\begin{equation}
SIOU_{when} = \frac{1}{|\mathcal{C}^+|} \sum_{c\in\mathcal{C}^+} \frac{\sum_t^{\mathcal{T}^+}{\hat{y}_t^c}}{\sum_t^{\mathcal{T}}{\hat{y}_t^c}},  
\end{equation}

\noindent where $\mathcal{C}^+$ and $\mathcal{T}^+$ are sets of true positive cameras and timesteps, respectively. $\hat{y}_t^c \in (0, 1)$ is the predicted value for the presence the object in camera $c$ at timestep $t$. Intuitively, $SIOU_{when}$ is a value between 0 and 1, representing the temporal overlap between the predicted and ground truth timesteps for all cameras where the individual appears. 

We compute the $SIOU$ for evaluating the \textit{where} problem as follows:

\begin{equation}
SIOU_{where} = \frac{1}{|\mathcal{C}^+|}  \sum_{c\in\mathcal{C}^+} \frac{1}{|\mathcal{T}^+|} \sum_t^{\mathcal{T}^+} \frac{\sum_g^{\mathcal{G}^+}{^g\hat{y}_t^c}}{\sum_g^{\mathcal{G}}{^g\hat{y}_t^c}},  
\end{equation}

\noindent where $\mathcal{G}^+$ is the set of grid cells where the individual is present. $SIOU_{where}$ is similarly a value between 0 and 1 representing the spatial overlap between predicted and ground truth grid cell locations for all true positive cameras and timesteps.

In addition to our new metrics, we also adapt the standard average displacement error (ADE) and final displacement error (FDE) metrics used widely in the SCTF literature \cite{behavior-cnn,social-lstm,social-gan,sophie}. We refer to these metrics as $ADE_{where}$ and $FDE_{where}$ hereinafter to distinguish them from their SCTF counterparts. We compute a single coordinate value for a heatmap $\mathboldit{H_i}$ by computing the center of mass, $R_{x,y}$, as follows:

\begin{equation}
R_{x,y} = ((\frac{1}{M}\sum_{m\in\mathboldit{H_i}}{x_m}\cdot r_x^m),((\frac{1}{M}\sum_{m\in\mathboldit{H_i}}{y_m}\cdot r_y^m))),
\end{equation}

\noindent where $M$ is the sum of all elements in $\mathboldit{H_i}$. Note that $\mathboldit{H_i}$ denotes a heatmap for camera $i$ at a single timestep. $r_x^m$ and $r_y^m$ respectively denote the $x$ and $y$ positions of the $m^{th}$ element in $\mathboldit{H_i}$. This value corresponds to the centroid of the target bounding box in the image space. As a target can appear in more than one camera view, we compute $ADE_{where}$ and $FDE_{where}$ separately for each camera the target appears and take the mean. $ADE_{where}$ and $FDE_{where}$ are valid for the \textit{where} problem formulation.

\section{Performance evaluation} \label{sec:evaluation}

In this section, we introduce the evaluation dataset, baseline approaches, and assess the performance of each model for the \textit{which}, \textit{when}, and \textit{where} tasks.

%
\subsection{Warwick-NTU multi-camera forecasting dataset}

We use the Warwick-NTU Multi-camera Forecasting (WNMF) dataset to train and evaluate each model, which is avaialbe to download at available to download from \url{https://github.com/olly-styles/Multi-Camera-Trajectory-Forecasting}. WNMF is collected specifically for MCTF using 15 cameras in a building on the Nanyang Technological University campus and contains both overlapping and non-overlapping views recorded over 20 days. The dataset consists of cross-camera trajectories where an individual departs from one camera view and then re-appears in another after no more than 12 seconds. An individual may also be visible in any number of other camera views during this tracking period. 

A single individual is labeled for each cross-camera trajectory, and interactions between individuals are limited due to low person density (1.41 individuals per camera per frame on average). Person bounding boxes are generated using pre-trained detection \cite{mask-rcnn} and tracking \cite{deepsort} algorithms and manually verified to be accurate. A human annotator verifies that each track consists of bounding boxes judged to have an IOU of $\geq$0.5 with the ground truth box. We do not provide identity labels; however, in practice, we observe that many cross-camera trajectories are unique identities due to the long period of data collection. We add two extensions to the dataset compared to our previous release: (i) Multi-viewpoint departures. The original dataset consists of matches across pairs of cameras. We add new annotations for overlapping viewpoints where an individual is visible in multiple cameras simultaneously. (ii) Cleaned annotations. The original dataset contains 2.3K cross-camera matches; however, several erroneous matches have been manually removed, leaving 2.0K matches. The erroneous matches are caused by high visual similarity between different persons during labeling. To verify cross-camera tracks, a human annotator reviews the track pairs proposed by the annotation procedure outlined in \cite{mctf} by referring to the camera network topology along with the appearance timestamps. Any tracks deemed to be false matches are discarded. More details on the dataset and semi-automated annotation procedure can be found in our previous work \cite{mctf}. We use a robust 5-fold cross-validation setup in all experiments that follow, where training and testing sets contain footage recorded on different days.

\subsection{Baseline approaches} \label{sec:baselines}

In addition to the models introduced in Section \ref{sec:models}, we also evaluate several baselines. 

\vspace{0.5mm}
\noindent\textbf{Shortest real-world distance.} We use the physical distance between cameras in the real world and predict the camera closest to the current camera. This baseline applies to the \textit{which} formulation only.

\vspace{0.5mm}
\noindent\textbf{Training set mean.} We consider all training set observations for a particular camera and take the mean of all ground truth labels in the training set.

\vspace{0.5mm}
\noindent\textbf{Most similar trajectory.} We find the most similar trajectory in terms of $L_2$ distance in the same camera from the training set to the observed trajectory and predict the same label.

\vspace{0.5mm}
\noindent\textbf{Hand-crafted features.} We extract some features from the bounding boxes and classify them with a fully-connected network. Our 10-dimensional hand-crafted feature vector contains velocity in $x$ and $y$ direction, acceleration in $x$ and $y$ direction, last observed bounding box height and width, and its four coordinates. We compute all features with respect to the 2D coordinate system as captured by the camera. 

\subsection{\textit{Which} camera will they appear?} \label{sec:which}

\textbf{Experimental setup.} We use our proposed network architectures (Fig. \ref{fig:mctf_models}) and baselines (Section \ref{sec:baselines}). For the coordinate trajectory approaches, we train a separate model for each camera. To adapt the SCTF models for MCTF, we leave encoders unchanged and change decoders to fully-connected output layers of size 15, the number of cameras in the WNMF dataset. We use a binary cross-entropy loss function with a sigmoid activation function at the output layer. The activation maps each output to a value between 0 and 1, representing the appearance of the individual in each camera. Coordinate trajectory and trajectory tensor models are trained with the Adam optimizer using learning rates of \num{1e-3} and \num{1e-4}, respectively. Coordinate trajectory encoders extract a feature vector of size 128, whereas trajectory tensor approaches extract a feature vector of size 512 as a single encoder is shared across all cameras and therefore requires larger representation capacity. All models are trained using a batch size of 64. We experiment with 3 sizes of input heatmap $\mathboldit{H}$: $16 \times 9$, $32 \times 18$, and $48 \times 27$. We use a Gaussian smoothing kernel size $\sigma$ between 0 and 4, chosen using cross-validation. The impact of changing these parameters is investigated in Section \ref{sec:additional}.

\begin{figure*}[t]
\centering
\includegraphics[width=\linewidth]{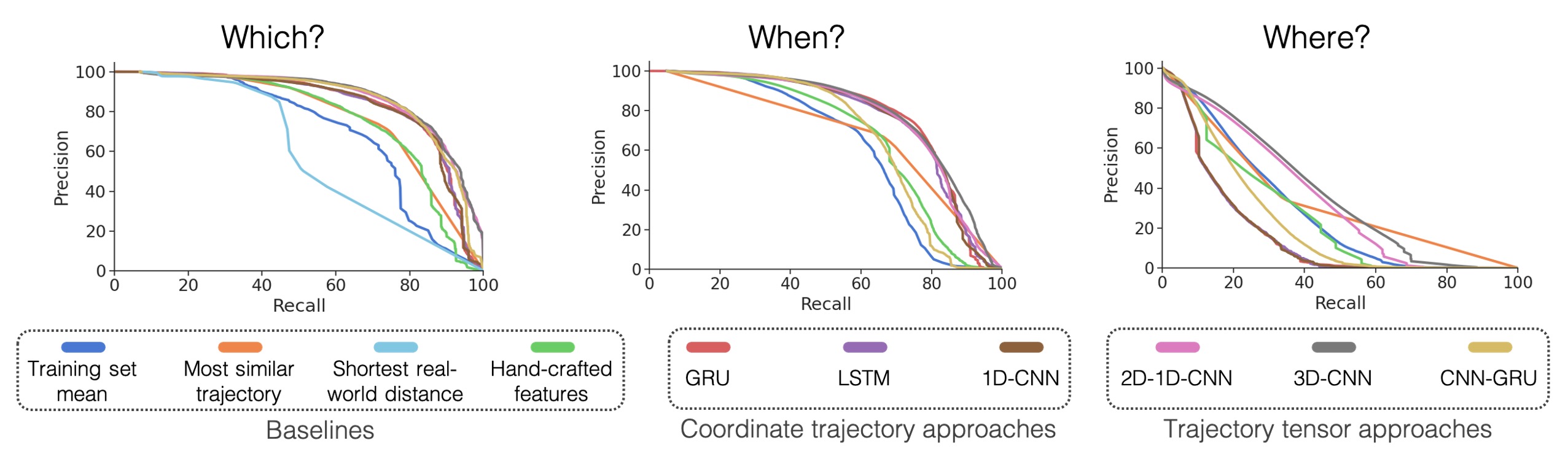}
\vspace{-4mm}
\caption{\textbf{Precision-recall plots.} Precision and recall of each model for all three problem formulations. Best viewed in colour.}
\label{fig:precision_recall}
\end{figure*}

\vspace{0.5mm}\noindent\textbf{Results.} The $AP_{which}$ for each model is shown in Table \ref{tab:which} and precision-recall plots are shown in Fig. \ref{fig:precision_recall} (left). Coordinate trajectory approaches outperform our 4 baselines, and trajectory tensor approaches perform the best.

\begin{table}[tb]
\setlength{\tabcolsep}{9pt} 
\begin{center}   \caption{\textbf{\textit{Which} results.} Given observations from one camera, each model predicts which camera(s) the person will re-appear.} 
\vspace{-3mm}
\renewcommand*{\arraystretch}{1.25}
  \begin{tabular}{l c c}
    \toprule
    \makecell{} & \bf Model & $AP_{which}$ (\textuparrow)\\ 
    \hline
    \multirow{4}{*}{Baselines} & Shortest real-world distance & 45.4\\
    & Training set mean & 68.9\\ 
    & Most similar trajectory & 65.3 \\
    & Hand-crafted features & 76.2 \\ 
    \noalign{\vskip 0.5mm}   \hdashline    \noalign{\vskip 0.5mm}    
    \multirow{3}{*}{\makecell{Coordinate \\ trajectories}} & LSTM & 84.1 \\
    & GRU &  84.0 \\ 
    & 1D-CNN & 83.3 \\ 
    \noalign{\vskip 0.5mm}    \hdashline    \noalign{\vskip 0.5mm}    
    \multirow{3}{*}{\makecell{Trajectory \\ tensors}} & 2D-1D-CNN & 87.1\\
    & \textbf{3D-CNN} & \textbf{87.5}\\
    & CNN-GRU & 86.1\\
    \bottomrule
    \vspace{-5mm}
  \end{tabular}\label{tab:which}
  \end{center}
  \end{table}

\subsection{\textit{When} will they appear?}
\textbf{Experimental setup.} We use the same encoders for evaluating \textit{when}, with either 1D-transposed convolutional layers or recurrent unit for decoding as shown in Fig. \ref{fig:mctf_models}. Each model is trained with the binary cross-entropy loss, and the trajectory tensor models are trained with the same hyperparameters as the \textit{which} problem formulation.

\vspace{0.5mm}\noindent\textbf{Results.} The $AP_{when}$ and $SIOU_{when}$ for each model are shown in Table \ref{tab:when}, and precision-recall plots are shown in Fig. \ref{fig:precision_recall} (center). We observe a similar trend to the results of the \textit{which} task, with the 3D-CNN trajectory tensor model performing best. Some baseline methods perform well in terms of $SIOU_{when}$, but poorly in terms of $AP_{when}$. This result suggests the most similar trajectory baseline effectively predicts the correct time window an individual will be visible in the target camera but often predicts the wrong camera.

\begin{table}[tb]
\setlength{\tabcolsep}{2pt} 
\begin{center} \caption{\textbf{\textit{When} results.} Given observations from one camera, each model predicts in which camera(s) the person will re-appear, and in which timesteps they will be present.} 
\vspace{-3mm}
\renewcommand*{\arraystretch}{1.25}
  \begin{tabular}{l c c c}
    \toprule
    \makecell{} & \bf Model & $AP_{when}$ (\textuparrow) & $SIOU_{when}$ (\textuparrow)\\ 
    \hline
    \multirow{3}{*}{Baselines} & Training set mean & 61.7 & 40.0\\
    & \textbf{Most similar trajectory} & 46.6 & \textbf{56.7}\\
    & Hand-crafted features & 67.1 & 48.7\\ 
    \noalign{\vskip 0.5mm}    \hdashline   \noalign{\vskip 0.5mm}    
    \multirow{3}{*}{\makecell{Coordinate \\ trajectories}} & LSTM & 76.8 & 51.8\\
    & GRU &  77.5 & 53.7\\ 
    & 1D-CNN & 77.1 & 52.8\\ 
    \noalign{\vskip 0.5mm}    \hdashline    \noalign{\vskip 0.5mm}    
    \multirow{3}{*}{\makecell{Trajectory \\ tensors}} & 2D-1D-CNN & 77.4 & 54.4\\
    & \textbf{3D-CNN} & \textbf{79.0} & 53.9\\
    & CNN-GRU & 69.1 & 42.0\\
    \bottomrule
    \vspace{-6.2mm}
  \end{tabular}\label{tab:when}
  \end{center}
  \end{table}

\subsection{\textit{Where} will they appear?}
\textbf{Experimental setup.} We use the same target heatmap size of $16 \times 9$ regardless of the input heatmap size for a fair comparison. We use this small heatmap size to reduce computational complexity, and $16 \times 9$ is the smallest possible while maintaining the original aspect ratio of the video frames. The output trajectory tensor is, therefore, of size $[15 \times 60 \times 16 \times 9]$. We do not apply Gaussian smoothing to the ground truth trajectory tensor targets. We use transposed convolutional layers to upsample extracted feature vectors to trajectory tensor outputs, representing an individual's future location. These are 3D transposed convolutions in the 3D-CNN model, and 2D transposed convolutions in the CNN-GRU and 2D-1D CNN models. The trajectory tensor models are trained with the same hyperparameters as in Section \ref{sec:which}, again using the binary cross-entropy loss.
\\
\noindent
\textbf{Results.} Results for each model are shown in Table \ref{tab:where} and precision-recall plots are shown in Fig. \ref{fig:precision_recall} (right). Note that due to the considerably increased complexity of the \textit{Where} problem formulation compared to the others, the AP is much lower. Results for this problem formulation are more mixed than for the \textit{Which} and \textit{When}, and some methods perform well on one metric but poorly on others. For example, the hand-crafted feature baseline performs the best in terms of $ADE_{where}$ and $FDE_{where}$, but poorly in terms of $AP_{where}$. Unlike the $AP_{where}$ metric, the displacement error and $SIOU_{where}$ metrics do not take into account erroneous predictions in camera views that the target does not appear. Therefore, this result suggests that the hand-crafted features baseline accurately forecasts the target trajectory but incorrectly assigns a high likelihood to the target appearing in other camera views.

\begin{table*}[tb]
\setlength{\tabcolsep}{6pt} 
\begin{center}   \caption{\textbf{\textit{Where} results.} Given observations from one camera, each model predicts which camera(s) the person will re-appear, in which timesteps they will be present, and where in the camera view they will appear. Best results are highlighted in \textbf{bold} typeface, second best are \underline{underlined}.} 
\vspace{-3mm}
\renewcommand*{\arraystretch}{1.25}
\begin{tabular}{l c c c c c}
    \toprule
    \makecell{} & \bf Model & $AP_{where}$ (\textuparrow)& $SIOU_{where}$ (\textuparrow)& $ADE_{where}$ (\textdownarrow)& $FDE_{where}$(\textdownarrow)\\ 
    \hline
    \multirow{3}{*}{Baselines} & Training set mean & 28.4 & 14.4 & 226.1 & \underline{260.4}\\
    & Most similar trajectory & 11.8 & \textbf{29.8} & 312.9 & 375.4\\
    & Hand-crafted features & 25.5 & 20.2 & \textbf{216.2} & \textbf{255.2}\\ 
    \noalign{\vskip 0.5mm}    \hdashline   \noalign{\vskip 0.5mm}    
    \multirow{3}{*}{\makecell{Coordinate \\ trajectories}} & LSTM & 16.3 & 8.7 & 285.9 & 365.5\\
    & GRU & 16.5 & 8.8 & 286.0 & 366.0\\ 
    & 1D-CNN & 16.4 & 8.6 & 287.2 & 368.1\\ 
    \noalign{\vskip 0.5mm}    \hdashline    \noalign{\vskip 0.5mm}    
    \multirow{3}{*}{\makecell{Trajectory \\ tensors}} & 2D-1D-CNN & \underline{34.5} & 22.6 & 225.9 & 265.8\\
    & 3D-CNN & \textbf{37.4} & \underline{22.9} & \underline{223.3} & 279.2\\
    & CNN-GRU & 22.4 & 12.8 & 280.1 & 351.7 \\
    \bottomrule
    \vspace{-7mm}
  \end{tabular}\label{tab:where}
  \end{center}
  \end{table*}

\subsection{Ablation studies} \label{sec:additional}

\vspace{0.5mm}\noindent\textbf{Multi-view trajectory tensors.} Our proposed trajectory tensor data representation enables us to model the same individual captured in multiple camera views. To study the impact of multiple camera views, we train our trajectory tensor-based models using only one of the available views, i.e., we set each heatmap at each timestep to the zero matrix for all but one of the camera channels $c \in \mathbf{Z}$. The results in Table \ref{tab:multi-view} show a comparison of our trajectory tensor models with a single-view trajectory (each heatmap is 0 for all but one of the cameras $c \in \mathbf{Z}$) and a multi-view trajectory (where more than one of the camera channels $c \in \mathbf{Z}$ may be non-zero). The results show that the models can exploit the location information available in multiple camera views.


\begin{table}[tb]
\setlength{\tabcolsep}{3pt} 
\begin{center}   \caption{\textbf{Multi-view trajectory tensor results.} Comparison of single-view and multi-view trajectories. Observing a trajectory in multiple camera views improves model performance for the \textit{Which} and \textit{Where} tasks.} 
\vspace{-2mm}
\renewcommand*{\arraystretch}{1.25}
  \begin{tabular}{l c c c c}
    \toprule
  \bf Model & \bf \makecell{Multi-view\\ trajectories} & $AP_{which}$ (\textuparrow) & $AP_{when}$ (\textuparrow) & $AP_{where}$(\textuparrow)  \\
    \hline
    \multirow{2}{*}{2D-1D-CNN}
    & No & 81.8 & 76.6 & \textbf{37.4}\\
    & Yes & \textbf{87.1} & \textbf{77.4} & 34.5\\
    \noalign{\vskip 0.3mm}
    \hdashline
    \noalign{\vskip 0.3mm}
    \multirow{2}{*}{3D-CNN}
    & No & 83.0 & 77.1 & \textbf{38.8}\\
    & Yes & \textbf{87.5} & \textbf{79.0} & 37.4\\
    \noalign{\vskip 0.3mm}
    \hdashline
    \noalign{\vskip 0.3mm}
    \multirow{2}{*}{CNN-GRU}
    & No & 83.4 & 68.4 & \textbf{23.6}\\
    & Yes & \textbf{86.1} & \textbf{69.1} & 22.4\\
    \bottomrule
  \end{tabular}\label{tab:multi-view}
  \vspace{-3mm}
  \end{center}
  \end{table}

\begin{figure}[t]
\centering
\includegraphics[width=0.9\linewidth]{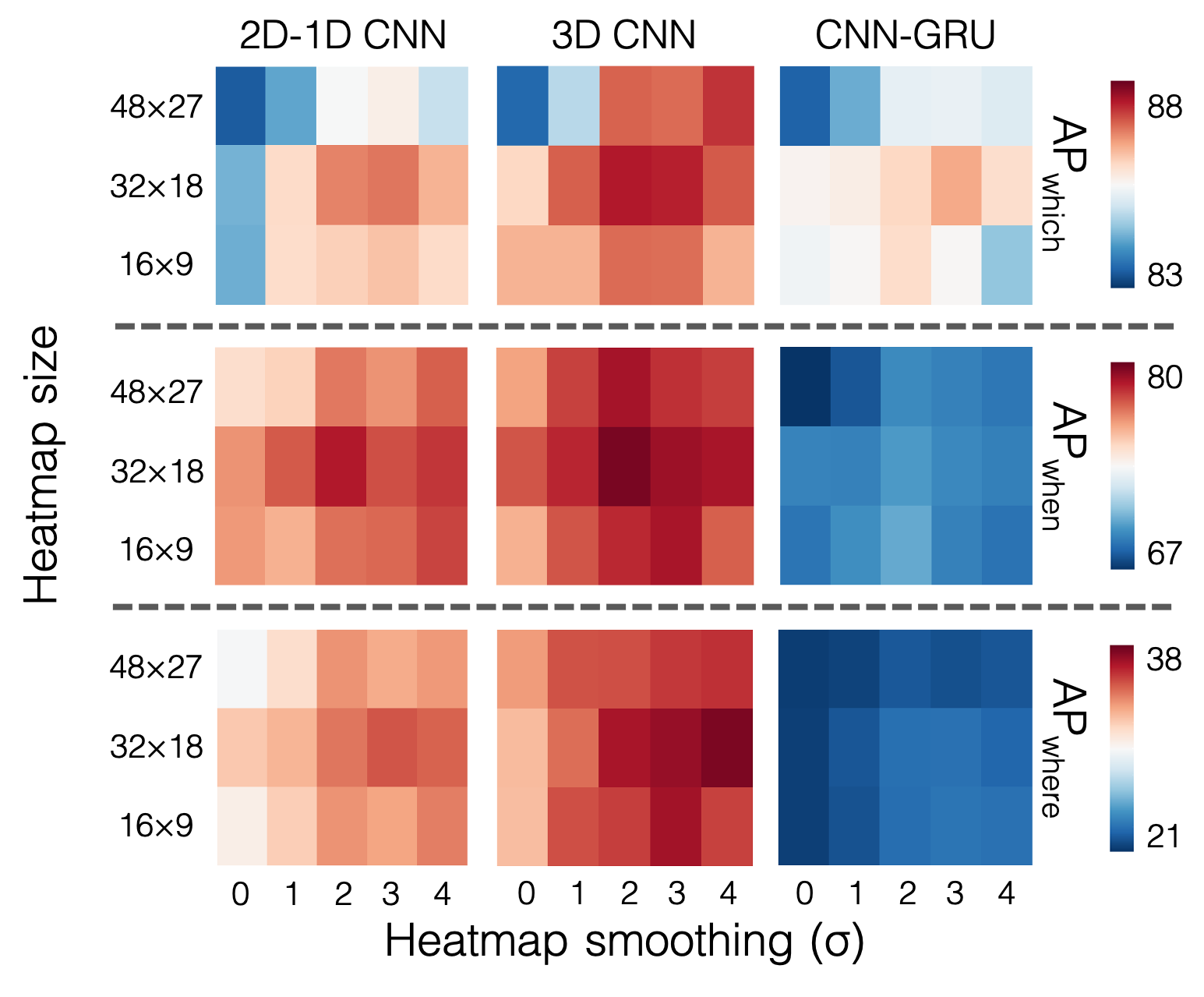}
\vspace{-4mm}
\caption{\textbf{Heatmap size and smoothing.} Impact on results of using different heatmap sizes and levels of smoothing ($\sigma$). Results shown are computed by 5 fold cross validation.}
\label{fig:ablation}
\end{figure}

\vspace{0.5mm}\noindent\textbf{Heatmap size and smoothing.} We investigate the impact of heatmap size and standard deviation of the Gaussian filter for smoothing. Larger heatmap sizes afford models more representation power at the cost of a larger number of parameters and a tendency to overfit the training data. On the other hand, smoothing has a regularizing effect by reducing the impact of errors during the detection and tracking stages. We suggest, therefore, that both heatmap size and smoothing $\sigma$ should be tuned in tandem. Fig. \ref{fig:ablation} shows the impact of heatmap size and smoothing $\sigma$. The results confirm our intuition that larger heatmap sizes generally benefit from more smoothing.

{\subsection{Multi-target multi-camera trajectory forecasting} \label{sec:mt-mctf}

Thus far, we have focused on the problem of single-target MCTF, where models predict the trajectory of a single target. We extend this to multi-target MCTF, allowing us to predict multiple target trajectories simultaneously. The WNMF dataset, as first introduced in our earlier work \cite{mctf}, does not contain multi-target MCTF labels, so we first generate a new set of annotations amenable to multi-target MCTF. 

\vspace{0.5mm}\noindent\textbf{Experimental setup.} We start our trajectory predictions when a target departs from a particular camera view. Therefore, for multi-target MCTF, we group trajectories into multi-target groups where each target departs at a similar time. Specifically, we group trajectories into bins of $2$ seconds ($10$ timesteps), such that targets with trajectories ending within $2$ seconds of each other are considered multi-target trajectories. As we predict the future trajectory using $2$ seconds of past data, grouping trajectories in this way ensures that trajectories overlap temporally for at least one timestep. Trajectories may be visible in the same or different camera views to other trajectories in the same multi-target group. To focus on the multi-target problem, we discard trajectories where only one target is visible, resulting in $58$ data samples, containing a total of $119$ trajectories, an average of $2.05$ trajectories per sample. The maximum number of trajectories in a single data sample is $4$. Due to the small size of the multi-target subset ($119$ trajectories compared to $1,967$ in the full dataset), results should be treated with caution. 

We evaluate our trained models and baselines on the multi-target subset of WNMF, which is available to download alongside the full dataset. We use $5$-fold cross-validation, using a model trained on data collected on different days to the test set. For our baselines and coordinate trajectory approaches, we process each trajectory sequentially. As our trajectory tensor-based approaches use a single unified model for all camera views, we process the multi-target trajectories in parallel by stacking trajectories along the batch dimension. Therefore, the input trajectory tensors are of dimension $b \times k \times t \times w \times h$, where $b$ is the number of targets, and other dimensions are the same as described in Section~\ref{sec:data_representation}. Stacking inputs along the batch dimension enables us to simultaneously predict trajectories for any arbitrary number of targets that fit into system memory. 

\vspace{0.5mm}\noindent\textbf{Results.} Results for multi-target MCTF are shown in Table~\ref{tab:multi-target}. Learned approaches outperform the baselines, although there is variation between the performance of the three tasks across models. This variation may be due to the small size of the multi-target subset compared to the full dataset. Our 2D-1D-CNN and 3D-CNN trajectory tensor models outperform coordinate trajectory approaches for the \textit{which} and \textit{where} tasks and process all targets and camera views using a single unified model rather than predicting trajectories for each target sequentially using a separate model for each camera.}

\begin{table}[tb]
\setlength{\tabcolsep}{2pt} 
\begin{center} \caption{\textbf{Multi-target multi-camera trajectory forecasting results.} Comparison of each model on a multi-target subset of the WNMF dataset. Trajectory tensor models stack multi-targets across an additional tensor dimension. Other models process each trajectory sequentially.}
\vspace{-3mm}
\renewcommand*{\arraystretch}{1.25}
  \begin{tabular}{l c c c c}
    \toprule
    \makecell{} & \bf Model & \makecell{$AP_{which}$ \\ (\textuparrow)} & \makecell{$AP_{when}$ \\ (\textuparrow)} & \makecell{$AP_{where}$ \\ (\textuparrow)}\\ 
    \hline
    \multirow{4}{*}{Baselines} 
    & Shortest real-world distance & 39.5 & N/A & N/A\\
    & Training set mean & 63.9 & 56.8 & 28.6\\
    & Most similar trajectory & 58.6 & 40.7 & 12.2\\
    & Hand-crafted features & 48.0 & 36.6 & 10.0\\ 
    \noalign{\vskip 0.5mm}    \hdashline   \noalign{\vskip 0.5mm}    
    \multirow{3}{*}{\makecell{Coordinate \\ trajectories}} & LSTM & 77.8 & \textbf{73.1} & 19.5\\
    & GRU & 73.5 & 72.3 & 20.7\\ 
    & 1D-CNN & 75.3 & 71.6 & 20.2\\ 
    \noalign{\vskip 0.5mm}    \hdashline    \noalign{\vskip 0.5mm}    
    \multirow{3}{*}{\makecell{Trajectory \\ tensors}} & 2D-1D-CNN & 70.8 & 70.6 & \textbf{33.4}\\
    & 3D-CNN & \textbf{80.0} & 63.3 & 30.2\\
    & CNN-GRU & 46.9 & 36.3 & 18.0\\
    \bottomrule
    \vspace{-6.2mm}
  \end{tabular}\label{tab:multi-target}
  \end{center}
  \end{table}

\subsection{Discussion and applications}
Our results show that learned models using either coordinate trajectories or trajectory tensors as inputs generally outperform other baselines at the three tasks in the MCTF framework. Representing trajectories as tensors allows us to model relative object locations in multiple camera views simultaneously, which generally improves MCTF performance when used as inputs and provides an intuitive way to represent multi-modal futures when used as targets. Trajectory tensors are also more elegant than using coordinate trajectories in an MCTF setting, as a single model can be used rather than creating separate models for each camera. Note that our approach and other baselines require that the camera network is the same at training and testing time.

We find that the 3D-CNN architecture consistently performs the best across MCTF tasks. 3D-CNNs have traditionally only seen notable success in settings where vast amounts of data are available for training \cite{can-3d-follow-2d}. However, in our setting, we use resolutions of up to $48 \times 27$ rather than the $224 \times 224$ or higher commonly used in activity recognition, which considerably reduces the number of parameters and facilitates training on smaller datasets. 

The benefits of the MCTF framework are fourfold: (i) Long-term forecasting. This is possible by removing the constraint of a single camera viewpoint. (ii) Intelligent camera monitoring. When tracking a particular object-of-interest across a camera network, a subset of cameras may be monitored intelligently using the predictions from an MCTF model rather than continually monitoring all cameras. (iii) Enhanced tracking. Location predictions may be used in conjunction with a RE-ID model for more robust multi-camera tracking. (iv) Robustness to camera failure. Predicting an individual's location in multiple camera views adds redundancy; i.e., a target may still be identified in the event that one or more cameras on the network are no longer operational.

\vspace{-2mm}

\section{Conclusion} \label{sec:conclusion}
 We have developed a complete framework for MCTF formulated in a hierarchy of three spatio-temporal localization tasks: In \textit{which} camera, \textit{when}, and \textit{where} will the object(s) appear? Our work is the first to address the challenges of trajectory forecasting in a multi-camera environment. We introduced the idea of the \emph{trajectory tensor} - a new trajectory representation that facilitates the encoding of multi-camera trajectories and associated uncertainty. Trajectory tensors are an attractive alternative to the traditional coordinate trajectory representation used in previous works. Our proposed MCTF models based on trajectory tensors show promising results on our WNMF database, and we hope to evaluate in more settings when more datasets for MCTF become available. We envision our MCTF model will enhance multi-camera surveillance systems, complementing existing models for person RE-ID and multi-camera tracking. 
 

\section*{Acknowledgments}

This work is funded by the UK EPSRC (grant number E/L016400/1). Our thanks to NVIDIA for their generous hardware donation. We would also like to thank the ROSE lab at the Nanyang Technological University, Singapore for providing the data used in this research.

\ifCLASSOPTIONcaptionsoff
  \newpage
\fi


\bibliographystyle{IEEEtran}
\bibliography{mybib}

\begin{IEEEbiography}[{\includegraphics[width=1in,height=1.5in,clip,keepaspectratio]{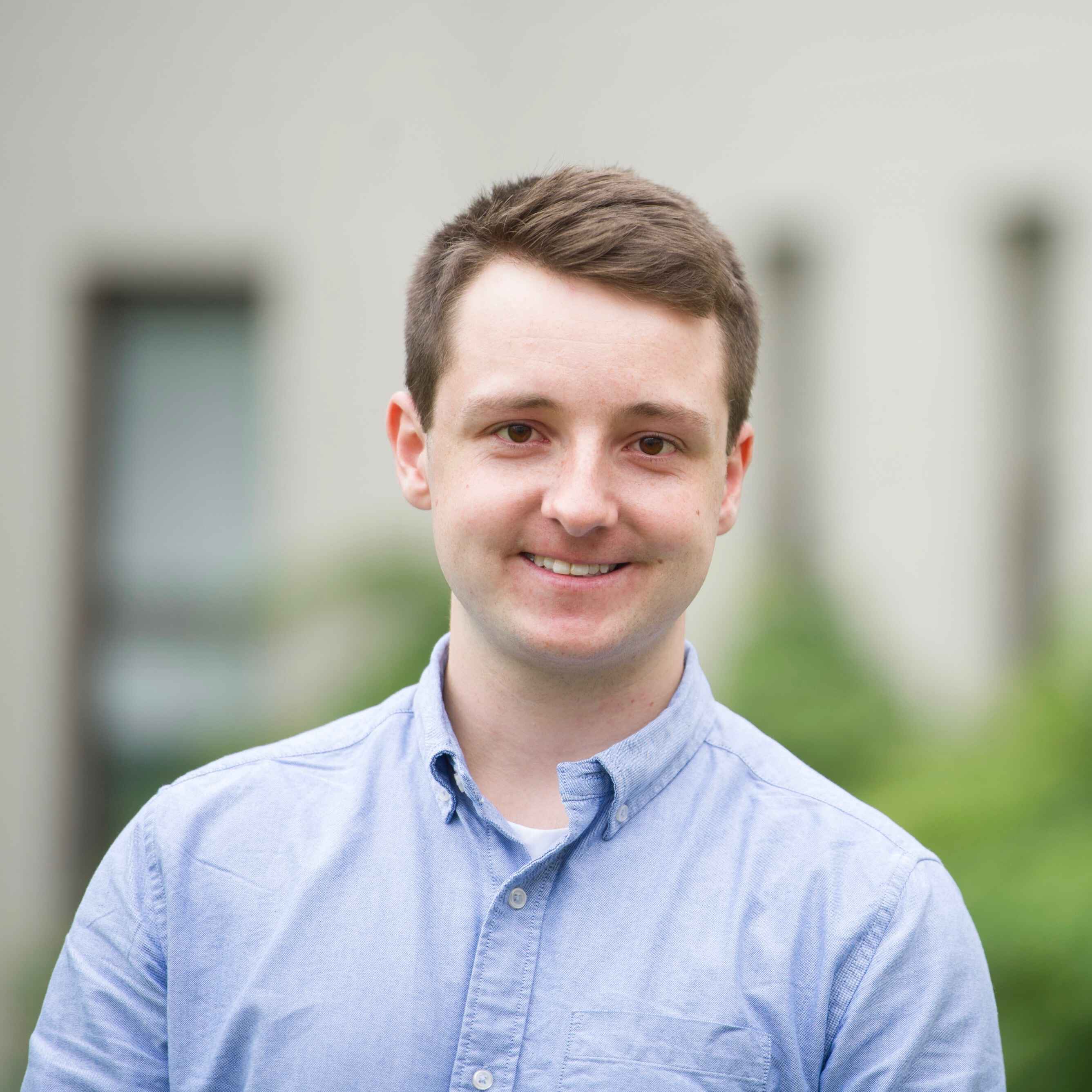}}]{Olly Styles}
is working towards a PhD degree at the University of Warwick, UK. He received an MEng degree in Computer Science, also from the University of Warwick. His research interests include vision-based forecasting, object tracking, and self-supervised learning. During his PhD, he has collaborated with several other research groups as a visiting scholar including the iProbe Lab at Michigan State University, USA, and the ROSE lab at Nanyang Technological University, Singapore.
\end{IEEEbiography}

\begin{IEEEbiography}[{\includegraphics[width=1in,height=1.5in,clip,keepaspectratio]{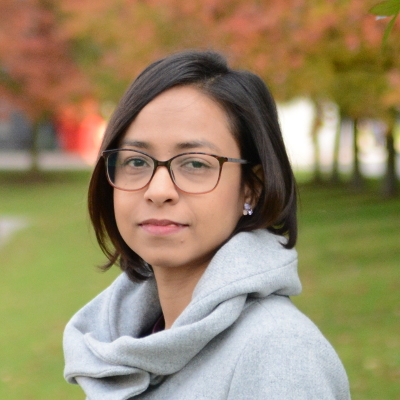}}]{Tanaya Guha}
is an Assistant Professor in the Department of Computer Science at the University of Warwick, UK. She has received her PhD in Electrical and Computer Engineering from the University of British Columbia (UBC), Vancouver, Canada. Her research is focused on modeling and analysis of multimedia data combining machine learning and signal processing with applications in media understanding, healthcare and smart surveillance. She has served in the Program Committee of several conferences including ACM Multimedia, IEEE ICME and Interspeech.
\end{IEEEbiography}

\begin{IEEEbiography}[{\includegraphics[width=1in,height=1.5in,clip,keepaspectratio]{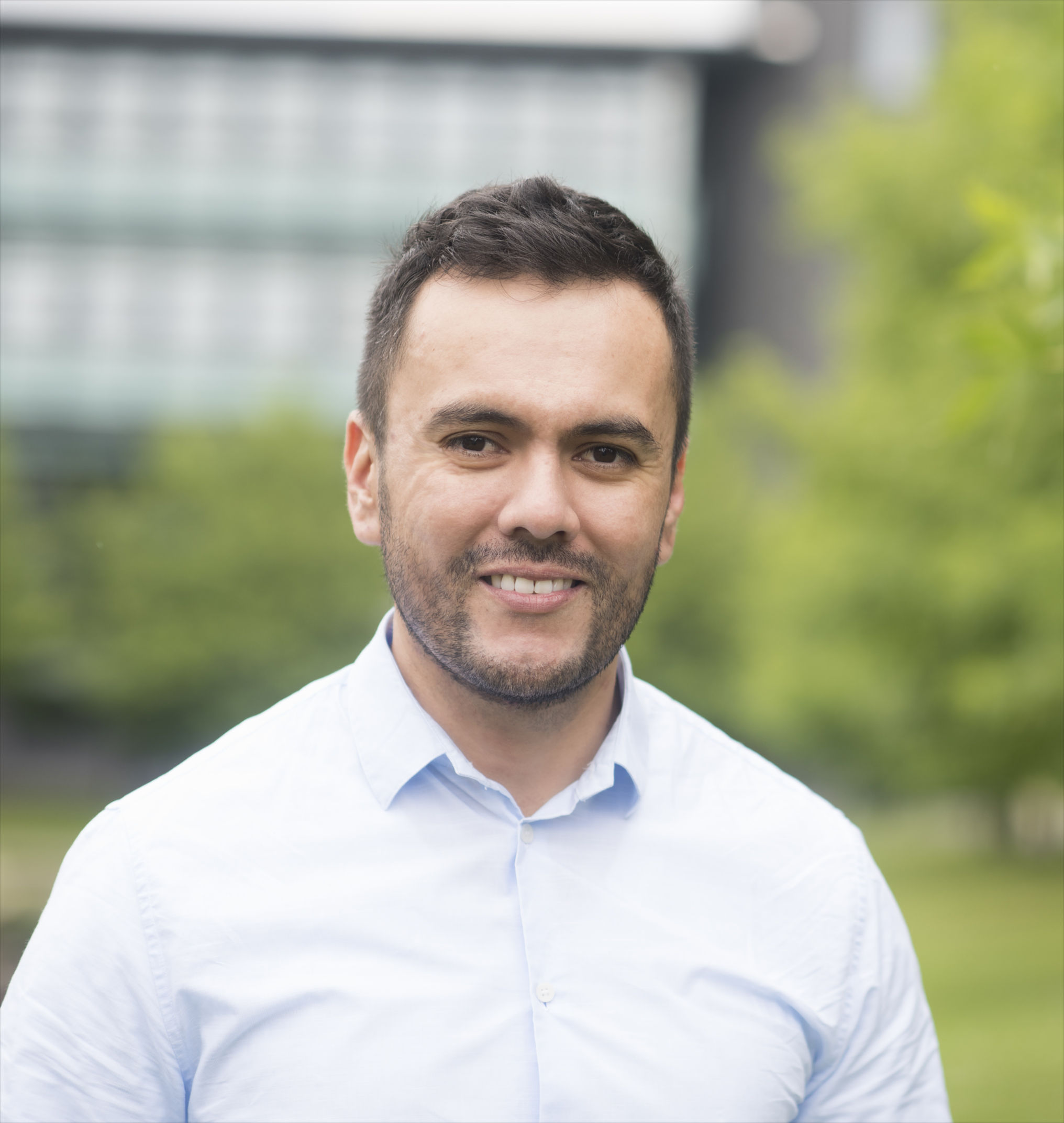}}]{Victor Sanchez}
received the MSc degree from The University of Alberta, Canada, in 2003, and the PhD degree from The University of British Columbia, Canada, in 2010. From 2011 to 2012, he was with the Video and Image Processing (VIP) Laboratory of The University of California at Berkeley as a Postdoctoral Researcher. In 2012, he was a Visiting Lecturer with the Group on Interactive Coding of Images, Universitat Autonoma de Barcelona. From 2018 to 2019, he was a Visiting Scholar with the School of Electrical and Information Engineering, The University of Sydney, Australia. He is currently an Associate Professor with the Department of Computer Science at The University of Warwick, U.K. His main research interests include signal and information processing and deep/machine learning with applications to multimedia data analysis, computer vision, image and video coding, security, and communications. He has authored several technical articles, book chapters, and a book in these areas. He is a member of the IEEE Inf. Forensics and Security Technical Committee, and IAPR-TC6, a Technical Committee on Computational Forensics under the auspices of the International Association for Pattern Recognition (IAPR). His research has been funded by the Consejo Nacional de Ciencia y Tecnologia, Mexico, the Natural Sciences and Engineering Research Council, Canada, the Canadian Institutes of Health Research, the FP7 and the H2020 Programs of the European Union, the Engineering and Physical Sciences Research Council, U.K., and the Defence and Security Accelerator, U.K.
\end{IEEEbiography}

\vfill

\end{document}